\documentclass[acmlarge]{acmart-ecaf}
\renewcommand{\baselinestretch}{1.15}
\setcopyright{cc}
\copyrightyear{2026}
\acmYear{2026}
\acmConference[ECAF'26]{Fifth European Conference on Algorithmic Fairness}{September 02--September 04, 2026}{Ghent, BE}
\usepackage{hyperref,url,graphicx,tikz,amsthm,rotating,stmaryrd,booktabs,siunitx}
\definecolor{XGBoostColor}{HTML}{348AA7}
\definecolor{GNNGINColor}{HTML}{5DD39E}
\definecolor{GNNPNAColor}{HTML}{BCE784}
\definecolor{IndirectEffect}{HTML}{138899}
\definecolor{DirectEffect}{HTML}{D75A4F}
\usetikzlibrary{positioning, arrows.meta}
\tikzset{>=Stealth,node distance=2cm,var/.style={font=\large\itshape},every path/.style={thick}}
\newcommand{\isdef}{\mathrel{\mathrel{\mathop:}=}}

\theoremstyle{definition}

\theoremstyle{remark}

\begin{document}
\title{Counterfactual Methods for Detecting Unfairness in Anti-Money Laundering Algorithms}
\author{Lea Multerer}
\affiliation{\institution{IDSIA}\city{Lugano}\country{Switzerland}}
\email{lea.multerer@idsia.ch}
\author{Michele Inchingolo}
\affiliation{\institution{IDSIA}\city{Lugano}\country{Switzerland}}
\email{michele.inchingolo@idsia.ch}
\author{David Kletz}
\affiliation{\institution{IDSIA}\city{Lugano}\country{Switzerland}}
\email{david.kletz@idsia.ch}
\author{Adrian Cosma}
\affiliation{\institution{IDSIA}\city{Lugano}\country{Switzerland}}
\email{adrian.cosma@idsia.ch}
\author{Alessandro Antonucci}
\affiliation{\institution{IDSIA}\city{Lugano}\country{Switzerland}}
\email{alessandro.antonucci@idsia.ch}
\author{Martina Gogova}
\affiliation{\institution{UBS Switzerland AG and its affiliates}\city{Frankfurt}\country{Germany}}
\email{martina.gogova@ubs.com}
\begin{abstract}
The application of machine learning–based predictive algorithms to \emph{Anti-Money Laundering} (AML) has grown rapidly, driven by the vast volume of financial transaction data available to banks. These algorithms are typically trained not only on transactional data but also on sensitive client information, which may raise fairness concerns. Despite this, AML detection systems remain largely underexplored from a fairness perspective, even though deeper analytical methods based on counterfactuals are now available. Such techniques enable the decomposition of the direct and indirect effects of potentially sensitive features on model predictions, thereby supporting the evaluation of whether their influence is acceptable from a fairness perspective. Closing this gap, we consider the synthetic \emph{IBM AMLSim} transaction dataset and construct additional features of the country of an account and its average behaviour. This improves the predictive performance of diverse machine learning models, ranging from baseline decision trees to state-of-the-art graph neural networks. We assess the potential unfairness associated with these features through a counterfactual, path-specific effect analysis. This reveals that fairness violations tend to be more pronounced for models whose predictive performance benefits the most from the extended features. Such a finding highlights a concrete instance of the trade-off between predictive accuracy and fairness in AML applications, thus underscoring the urgency of a systematic fairness analysis in such critical domains.
\end{abstract}
\maketitle
\section{Introduction}\label{sec:intro}
\emph{Anti-Money Laundering} (AML) refers to the set of practices aimed at preventing the movement of illicit funds through the financial system. The responsibility for detecting such activities primarily lies with financial institutions, which are expected to implement \emph{Know Your Customer} (KYC) standards, monitor transactions, suspend accounts deemed suspicious, and submit timely reports to regulators. To address these challenging tasks, banks can leverage the large volumes of transaction data they possess, naturally motivating the adoption of machine learning algorithms to, at least partially, automate AML processes.

From a machine learning perspective, the historical data available to banks can be viewed as a supervised dataset, involving descriptive features for the transactions together with a label indicating whether a transaction is potentially suspicious, thereby defining a binary target variable. While training a classifier on such data is straightforward, achieving satisfactory predictive performance may be challenging for several reasons. First, the annotations may be noisy or incomplete, particularly for transactions labelled as negative (i.e., non-suspicious). Second, the dataset is typically highly imbalanced, as potential money laundering cases constitute only a small fraction of all transactions~\cite{nazar_magnitude_2024}. Finally, money laundering strategies are heterogeneous (e.g., circular or repeated transfers among the same counterparties, or other patterns that obscure the origin and destination of funds) and evolve over time.

AML regulatory frameworks and supervisory guidance explicitly encourage the use of specific risk indicators, such as jurisdictional information, customer residency status, account age, and cross-border transaction patterns. These indicators are typically captured as part of a KYC process.
This refers to the information used by banks to build a basic profile of their customers before providing services, as laid out in international standards and implemented through national regulation. These KYC-related, potentially sensitive, or \emph{protected}, features are routinely incorporated into risk-based AML programs and international guidelines. For instance, in Switzerland, these KYC obligations are defined by the \emph{Anti‑Money Laundering Act}\footnote{See \href{https://www.fedlex.admin.ch/eli/cc/1998/892_892_892/en}{fedlex.admin.ch/eli/cc/1998/892\_892\_892/en}.} (AMLA) and related ordinances. At the same time, recent regulatory and policy developments increasingly emphasise that machine learning-based decision systems should be transparent, explainable, and avoid unjustified reliance on protected attributes. This creates a structural tension between regulatory legitimacy, which requires the operationalsation of prescribed risk factors for financial crime prevention, and the need to assess the fairness of predictive algorithms, particularly with respect to whether automated decisions depend on such attributes beyond what can be justified by task-relevant behaviour.

This tension naturally motivates the use of \emph{algorithmic fairness}, which has emerged as a principled framework for evaluating bias in machine learning models~\cite{kusner_counterfactual_2017,plecko_causal_2024} and has found applications across various domains. As expected, much of the literature on machine learning approaches to AML focuses primarily on predictive performance~\cite{altman_realistic_2023,motie_financial_2024,gu_optimization_2025}. Fairness analyses in AML remain limited and are predominantly based on correlational approaches~\cite{kamalaruban_evaluating_2024,mazumder_explainable_2026}. To the best of our knowledge, our work is the first to adopt a causal perspective, grounded in mediation analysis~\cite{pearl_causality_2009}, to make these tensions in AML explicit. We study the potential impact of a protected feature through a counterfactual perspective~\cite{chiappa_path-specific_2019}. This enables a more granular assessment of how protected attributes affect model decisions through distinct causal pathways.

Common situations of interest are related to the presence of compound risk indicators explicitly referenced in international AML guidance, such as red flags issued by the \emph{Financial Action Task Force}\footnote{See \href{https://www.fatf-gafi.org}{fatf-gafi.org}.} (FATF). Examples include the combination of newly opened accounts and non‑domestic customer status, which is commonly associated with elevated money‑laundering risk at the population level. When operationalised in automated monitoring systems, however, such indicators can systematically affect specific customer segments, such as international students or temporary residents, who may legitimately exhibit short account histories and cross‑border transaction patterns. As a result, these groups may be disproportionately flagged not due to illicit behaviour, but because they structurally align with multiple regulatory risk factors. Our counterfactual and path‑specific analysis is therefore not intended to challenge the validity of FATF guidance, but to make explicit whether model predictions rely primarily on behavioural mediators or on direct dependence on sensitive characteristics. To make these ideas more explicit, let us consider the following simplified situation:

\begin{quote}
{\it A bank uses a machine learning model to flag transactions for AML investigation. Customers differ in a protected demographic attribute, and also in behavioural risk features derived from their transaction patterns. Customers from one demographic group (group~A) are flagged more frequently than those from another group (group~B). Part of this may be explained by differences in these behavioural risk features, which are associated with higher money-laundering risk and are therefore treated as legitimate risk indicators by the bank. However, even for customers with similar behavioural profiles, individuals from group~A are more likely to be flagged, indicating a direct dependence on the protected attribute.} 
\end{quote}

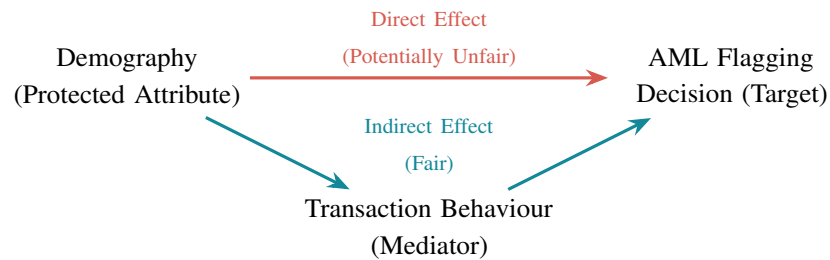
\begin{figure}[htb!]
\centering
\caption{Causal mediation example for AML flagging decisions. The protected demographic attribute influences the probability of a transaction being flagged both indirectly through behavioural features and directly, capturing the potentially unfair component not explained by differences in behaviour.}\label{fig:example_mediation}
\Description{Causal diagram with three nodes: Demography (Protected Attribute), Transaction Behaviour (Mediator),
and AML Flagging Decision (Target). Arrows go from Demography to Transaction Behaviour, from Transaction Behaviour to AML Flagging Decision, and directly from Demography to AML Flagging Decision. The path through Transaction Behaviour represents the indirect, potentially legitimate effect, while the direct arrow from Demography to AML Flagging Decision represents the potentially unfair direct effect of the protected attribute.}
\begin{tikzpicture}[scale=1]
\node[text centered,text width=30mm] at (0, 0) (a) {Demography\\(Protected Attribute)};
\node[text centered,text width=33mm] at (4,-2) (b) {Transaction Behaviour\\(Mediator)};
\node[text centered,text width=30mm] at (8,0) (c) {AML Flagging \\Decision (Target)};
\draw [->, very thick, color=IndirectEffect] (a) to (b);
\draw [->, very thick, color=DirectEffect] (a) -- node[midway,above,text centered,text width=30mm] {\footnotesize Direct Effect\\ (Potentially Unfair)}  (c);
\draw [->, very thick, color=IndirectEffect] (b) to (c);
\node[text centered, color=IndirectEffect ,text width=30mm] at (4,-0.9) (b) {\footnotesize Indirect Effect\\(Fair)};
\end{tikzpicture}
\end{figure}

A causal perspective allows the total effect of the protected attribute on the decision to be decomposed into a component transmitted via risk-relevant mediators, which can be acceptable from a fairness perspective, and a component that reflects a direct dependence on the protected attribute, to be instead considered unacceptable. Figure~\ref{fig:example_mediation} schematically represents such a decomposition.

In this work, we formalise a fairness
auditing framework for AML transaction monitoring systems within a structural causal model. We evaluate different classification algorithms, ranging from relatively simple ensembles of decision trees (e.g., XGBoost), which exhibit comparatively lower predictive performance, to more sophisticated graph neural networks achieving state-of-the-art results.
Due to the confidential nature of transactions and AML-related annotations, most academic machine learning studies in this domain rely on synthetic datasets. In line with this practice, we consider the \textit{IBM AMLSim} datasets, which are synthetically generated to reflect realistic AML patterns~\cite{altman_realistic_2023,egressy_provably_2024}. However, these datasets do not include KYC features. To enable an analysis of algorithmic fairness issues, we introduce \emph{pseudo} KYC features corresponding to the country of the sending account of each transaction. We treat this categorical feature as a protected attribute and consider, as a potential mediator of the effect on the target variable (i.e., whether a transaction is potentially suspicious), a numerical feature derived from the historical transaction profiles of each customer. We find that the inclusion of our pseudo KYC features leads to improved predictive performance and fairness violations tend to be more pronounced for models that benefit the most from these features. This is in line with the expected accuracy-fairness trade-off~\cite{plecko_fairness-accuracy_2025}.

The paper is organised as follows. In Section~\ref{sec:background}, we quickly review the basic notions of the counterfactual approach to algorithmic fairness. In Section~\ref{sec:data}, we introduce the dataset used for the analysis and our design choices to extract the descriptive features. The inference algorithms we consider and their performance on the datasets are described in Section~\ref{sec:algos}. The results of our fairness analysis are in Section~\ref{sec:fairness}. Conclusions and limitations are in Section~\ref{sec:discussion}.
\section{Background on Counterfactual Fairness Analysis}\label{sec:background}
We begin by introducing the necessary notation. Uppercase letters denote random variables, lowercase letters denote their states, and calligraphic uppercase letters denote the corresponding state spaces. Accordingly, $v \in \mathcal{V}$ represents a state of the variable $V$. Given a distribution over $V$, denoted by $P(V)$, the expectation of a function $f$ of $V$ is defined as:
\begin{equation}
\mathbb{E}[f] \isdef \sum_{v \in \mathcal{V}} P(v) \cdot f(v)\,,
\end{equation}
with the sum replaced by an integral in the case of continuous variables. Finally, Iverson brackets $\llbracket \cdot \rrbracket$ denote an indicator function that is one if its argument is true and zero otherwise. Note that, for instance, $\mathbb{E}[\llbracket V=v \rrbracket]=P(v)$.

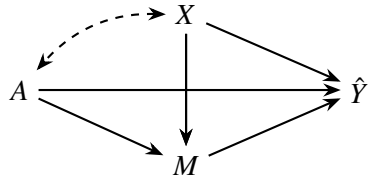
\begin{figure}[htp!]
\centering
\caption{Structural causal graph with protected attribute $A$, mediators $M$, covariates $X$, and classifier prediction $\hat Y$.}\label{fig:scm}
\Description{Directed acyclic graph with nodes $A$, $X$, $M$, and $\hat Y$; arrows from $A$ and $X$ to $M$, and from $A$, $X$, and $M$ to $\hat Y$.}
\begin{tikzpicture}
\node[var] (A) at (-1.5,0) {$A$};
\node[var] (Y) at ( 3,0) {$\hat{Y}$};
\node[var] (X) at ( 0.7,1) {$X$};
\node[var] (M) at ( 0.7,-1) {$M$};
\draw[->] (A) -- (Y);
\draw[->] (A) -- (M);
\draw[->] (X) -- (Y);
\draw[->] (X) -- (M);
\draw[->] (M) -- (Y);
\draw[->] (X) -- (M);
\draw[dashed,<->,bend right=25] (X) to (A);
\end{tikzpicture}
\end{figure}

We consider a classification task with target variable $\hat{Y}$, denoting the prediction returned by a classifier. Let $A$ denote one of the relevant features for predicting $\hat{Y}$, which we treat as a \emph{protected} attribute that may raise fairness concerns. Our goal is to evaluate the impact of the input attribute $A$ on the output $\hat{Y}$ from a fairness perspective, using tools from causal inference~\cite{pearl_causality_2009,chiappa_path-specific_2019}. This requires assumptions about the underlying causal relationships among $A$, $\hat{Y}$, and the other relevant features used by the classifier. Since $\hat{Y}$ is, by construction, a deterministic function of the input features, all such features have a direct impact on $\hat{Y}$. We define an attribute as a \emph{mediator} between $A$ and $\hat{Y}$ if, in addition to having a direct effect on $\hat{Y}$, it is also directly influenced by $A$, thereby transmitting part of the effect of $A$ to $\hat{Y}$. We denote the set of mediators by $M$ and the remaining features by $X$. These causal relationships can be naturally represented as edges in a directed graph, yielding the topology shown in Figure~\ref{fig:scm}, also referred to as the \emph{standard fairness model}~\cite{plecko_causal_2024}.

The distinction between the mediators in $M$ and the remaining attributes $X$ can be established based on domain expertise; alternatively, dedicated knowledge discovery algorithms may be employed. The directed, acyclic graph in Figure~\ref{fig:scm} is the qualitative basis for defining a \emph{structural causal model} over the (\emph{endogenous}) variables $(A,X,M,\hat{Y})$ and the auxiliary, \emph{exogenous}, variables $U_{A,X}$ and $U_M$, adding stochasticity to the model. The endogenous variables are determined by structural equations having as input the variables corresponding to the \emph{parent} nodes in the graph together with the corresponding exogenous variables, i.e.,
\[
A \isdef f_A(U_{A,X}),\quad
X \isdef f_X(U_{A,X}),\quad
M \isdef f_M(A,X,U_M),\quad
\hat{Y} \isdef f_{\hat{Y}}(A,X,M)\,.
\]

Note that the equation $f_{\hat{Y}}$ is the deterministic classifier we learn from the data, and for this reason it does not include the dependency of any exogenous variable. The other endogenous variables, instead, depend on exogenous variables which are typically unobservable, preventing us from specifying analogous deterministic functions. Note also that $U_{A,X}$ appears in the equations of both $A$ and $X$, thus acting as an exogenous \emph{confounder} between these endogenous variables. In causal graphs such as the one in Figure~\ref{fig:scm}, confounders are expressed by bidirected dashed arcs, while exogenous variables affecting a single endogenous variable are omitted.

An \emph{intervention} on an endogenous variable of a structural causal model describes how the system would behave in a hypothetical scenario where that variable is externally set to a particular value, while all other causal mechanisms remain unchanged. For the protected variable $A$, this corresponds to replacing the structural equation for $A$ by the constant $A=a$, while leaving all other structural equations unchanged, i.e.,
\[
A = a,\quad
X = f_X(U_{A,X}),\quad
M_{A \leftarrow a} \isdef f_M\big(a, X, U_M\big), \quad
\hat Y_{A \leftarrow a} \isdef
f_{\hat Y}\big(a, X, M_{A \leftarrow a}\big)\,.
\]

\emph{Counterfactual fairness} quantifies how sensitive the predictions of a classifier are to interventions forcing the protected attribute $A$ to a state different from the observed one or from the one used as reference. A natural first diagnostic of this sensitivity is the counterfactual \emph{flip rate} with respect to a reference value of the protected attribute $A$ denoted by $a'\in\mathcal{A}$. It quantifies how often the prediction changes when $A$ is set to this reference state, i.e.,
\begin{equation}\label{eq:FR}
\mathrm{FR}_{\hat Y}(a')
\isdef
\mathbb{E}\big[\llbracket \hat Y \neq \hat Y_{A \leftarrow a'}\rrbracket\big]\,,
\end{equation}
that is, the probability that the prediction would change if the protected attribute of a transaction were set to the reference value $a'$.

While the flip rate captures the overall sensitivity of predictions to interventions on $A$, it does not reveal along which causal pathways this sensitivity arises. To disentangle fair from unfair influences of $A$, the effect of $A$ on $\hat Y$ is examined through the mediator $M$. In typical applications (see also the example in Section~\ref{sec:intro}), the mediated path $A \rightarrow M \rightarrow \hat Y$ is regarded as legitimate, whereas the direct path $A \rightarrow \hat Y$ is regarded as unfair.

Given two states $a,a' \in \mathcal{A}$ of the protected attribute, the \emph{total effect} \cite{plecko_causal_2024} of changing $A$ from $a'$ to $a$ is defined as:
\begin{equation}\label{eq:TE}
\mathrm{TE}_{\hat Y}(a,a')
\isdef \mathbb{E}\big[\hat Y_{A \leftarrow a}\big] - \mathbb{E}\big[\hat Y_{A \leftarrow a'}\big]\,.
\end{equation}
Let $M_{A \leftarrow a}$ and $M_{A \leftarrow a'}$ denote the mediator under interventions $A{=}a$ and $A{=}a'$, respectively.
The total effect $\mathrm{TE}_{\hat Y}(a,a')$ can be decomposed into the sum of the \emph{natural direct} ($\mathrm{NDE}$) and \emph{natural indirect} ($\mathrm{NIE}$) effects, i.e.,
\[ \mathrm{TE}_{\hat Y}(a,a') = \mathrm{NIE}_{\hat Y}(a,a') + \mathrm{NDE}_{\hat Y}(a,a')\,,\]
where
\begin{equation}\label{eq:NIE}
\mathrm{NIE}_{\hat Y}(a,a')
\isdef \mathbb{E}\big[\hat Y_{A \leftarrow a,\,M \leftarrow M_{A \leftarrow a}}\big]
- \mathbb{E}\big[\hat Y_{A \leftarrow a,\,M \leftarrow M_{A \leftarrow a'}}\big]
\end{equation}
captures the effect transmitted along the mediated path $A \to M \to \hat Y$, and
\begin{equation}\label{eq:NDE}
\mathrm{NDE}_{\hat Y}(a,a') \isdef \mathbb{E}\big[\hat Y_{A \leftarrow a,\,M \leftarrow M_{A \leftarrow a'}}\big] - \mathbb{E}\big[\hat Y_{A \leftarrow a',\,M \leftarrow M_{A \leftarrow a'}}\big]
\end{equation}
captures the contribution of the direct path $A \to \hat Y$, with the mediator path held at the baseline level corresponding to $A{=}a'$.

The quantities in Eqs.~\eqref{eq:TE}, \eqref{eq:NIE}, and \eqref{eq:NDE} will be used for our fairness analysis. If the total effect is not negligible, we derive its direct and indirect components. We typically regard a non-negligible direct effect as unacceptable from a fairness perspective, while possibly considering acceptable an indirect effect obtained through mediator variables.
\section{Synthetic AML Transaction Data}\label{sec:data}
In this section, we introduce the experimental setup and provide details about the synthetic data and their features for reproducibility. We base our study on synthetic transaction datasets for AML~\cite{altman_realistic_2023, egressy_provably_2024}. These datasets consist of realistic financial transactions simulated in a virtual economy of individuals, companies, and banks, where a small subset engage in illicit activities and launder funds through multi-step transactions. They are designed to support the development and evaluation of AML machine-learning detection models. Two groups of three datasets, each one with a different imbalance ratio, are available. For our preliminary study, we use the \emph{HI-Small} dataset, which is the smallest one among the group with the highest rate of money laundering activities.

The dataset contains $N_T$ = 5\,078\,345 transactions spanning 18 days. The transactions involve $N_A$ = 515\,088 accounts and 30\,470 unique banks. Each transaction record includes a timestamp, identifiers of the sending and receiving institutions with account numbers, amounts and currencies exchanged, payment format, and a Boolean ground-truth label stating whether or not the transaction was potentially suspicious from an AML perspective. The account data contain identifiers linking the accounts to entities and banks. The prevalence of transactions labelled as positive is around 0.1\%, this corresponding to 5\,177 transactions, with 6\,357 accounts involved. These levels align with the values traditionally observed in real AML monitoring activities (often around 1\%).

To perform a counterfactual fairness analysis on such data, features that serve as a protected attribute and a mediator are constructed from the available account information. We consider the \emph{country} associated with each account as protected attribute. An important remark is that, as we work with synthetic data, the names of the countries are reported just for illustrative reasons: the results do not reflect country-specific patterns. Lists of high-risk countries\footnote{See, e.g.,
\href{https://finance.ec.europa.eu/financial-crime/anti-money-laundering-and-countering-financing-terrorism-international-level_en}{finance.ec.europa.eu/financial-crime/anti-money-laundering-and-countering-financing-terrorism-international-level\_en}.} are routinely compiled by regulators. Biases towards the countries in these lists are expected. Notably, the countries involved in our data are not included in these lists.  Country information is not directly available in the dataset, and we retrieve it through a rule-based mapping based on string-matching patterns and processing of the bank names. For bank names that cannot be matched to a specific country, we consider the transaction currencies. All outgoing transactions from the bank are identified, and the country of the most frequent currency is assigned. Overall, this defines a categorical variable with 34 possible values, also including a dedicated value for transactions involving \emph{crypto} currencies. The features associated with the sending and receiving countries are attached to the transaction dataset. 

We conjecture that the country of an account influences both the frequency and the volume of transactions, reflecting differences in payment habits as well as country-specific costs. We therefore construct an additional feature in this simulated dataset that can serve as a mediating variable. To this end, we calculate the total number of transactions and the transaction volume over the data window for each account. Transaction volume is represented in units of 1\,000 US dollars, using a fixed table of exchange rates. The \emph{transaction behaviour} of an account is then defined as the sum between the $\log(\text{Number of transactions} + 1)$ and $\log(\text{Transaction volume in 1\,000 USD} + 1)$, yielding a single scalar measure per account that increases with higher activity, larger monetary flow, or both, while reducing the influence of extreme accounts. The mean transaction behaviour of the accounts, summarized at the sending country level, is displayed in Table~\ref{tab:summary_transactions} together with the number of transactions. Our transaction behaviour feature is attached to the transaction data for both sending and receiving accounts. The table also depicts the, variable, rate of suspicious activities for the different countries.

\begin{table}[htb!]
\caption{Summary of the transaction data categorized by sending country. Results are based on simulated data and are not expected to reflect country-specific patterns. Only countries with more than 30\,000 transactions are displayed.}
\label{tab:summary_transactions}
\begin{tabular}{lrrr}
\toprule
\textbf{Country} & \textbf{Transactions} ($N_T$) & \textbf{Mean Transaction Behaviour} & \textbf{Laundering Events}  [\%]\\
\midrule
United States & 2\,295\,079 & 13.69 & 0.10 \\
Germany & 320\,556 & 10.26 & 0.10 \\
China & 261\,709 & 10.63 & 0.11 \\
Switzerland & 220\,803 & 10.14 & 0.08 \\
India & 178\,466 & 10.15 & 0.07 \\
United Kingdom & 170\,525 & 10.20 & 0.09 \\
France & 163\,661 & 9.68 & 0.10 \\
Japan & 162\,993 & 10.08 & 0.10 \\
Israel & 160\,005 & 9.95 & 0.08 \\
Italy & 133\,910 & 9.74 & 0.09 \\
Australia & 118\,663 & 10.06 & 0.09 \\
Canada & 117\,691 & 10.04 & 0.08 \\
Spain & 111\,538 & 9.73 & 0.14 \\
Russia & 102\,348 & 9.98 & 0.08 \\
Mexico & 76\,608 & 9.96 & 0.08 \\
Saudi Arabia & 62\,265 & 10.02 & 0.45 \\
Brazil & 49\,046 & 10.20 & 0.10 \\
Netherlands & 47\,687 & 9.42 & 0.10 \\
Belgium & 41\,134 & 9.68 & 0.12 \\
Austria & 38\,534 & 9.91 & 0.11 \\
Finland & 35\,463 & 11.36 & 0.24 \\
Portugal & 31\,084 & 10.14 & 0.11 \\
\bottomrule
\end{tabular}
\end{table}

This leads to two versions of transaction data. The basic dataset, denoted henceforth as $T_B$, contains only the original transaction attributes of \emph{HI-Small}. The extended dataset, denoted by $T_E$, augments these attributes with the sender and receiver country of the accounts and mediator features, a transaction behaviour feature. To assess how strongly the transaction behaviour is tied to the country of origin, we perform a least squares fit with country being a fixed effect. The model explains about 9\% of the variance in the mediator ($R^2 =$ 0.09), relative to an intercept-only model. Hence, we conclude that the constructed transaction behaviour does capture country-specific transactional behaviour, but most of its variability still occurs within countries. This is consistent with its role as a soft, behaviour-based mediator.
\section{AML Classifiers}\label{sec:algos}
We now turn to machine learning classifiers and their performance for detecting money laundering activities in the datasets $T_B$ and $T_E$. Detecting money laundering on such highly imbalanced and relational data is known to be challenging, and substantial work has already focused on improving predictive performance for AML~\cite{altman_realistic_2023,egressy_provably_2024}. In this study, we aim to analyse counterfactual fairness properties of representative models. We therefore adopt classifier architectures that are standard in the AML and graph-learning literature rather than tuning models for maximal performance. As a tabular baseline, we apply a standard gradient-boosted decision tree model (XGBoost), which is expected to have limited performance. Nevertheless, such an implementation provides a realistic benchmark for what a single financial institution, training only on its own transactional data, might deploy in practice. Substantially higher performance would be difficult to achieve without access to richer, cross-institutional information. As transactions implicitly define a network structure, \emph{Graph Neural Networks} (GNN) now serve as state-of-the-art for financial fraud detection~\cite{motie_financial_2024}. For this reason, we consider two GNN implementations following the open-source implementation from the original benchmark publication~\cite{altman_realistic_2023,egressy_provably_2024}. As a GNN baseline, a \emph{Graph Isomorphism Network} (GNN-GIN) is used, and a \emph{Principal Neighbourhood Aggregation network} (GNN-PNA) is used as the best-performing model in the original publication~\cite{altman_realistic_2023,egressy_provably_2024}. For this study, GNN-GIN and GNN-PNA provide representative, state-of-the-art graph-based classifiers on which to study counterfactual fairness in AML.

The two model types operate on the same set of transactions but using different input representations. XGBoost receives a flat tabular feature vector, whereas the GNNs encode bank and account identities as graph nodes and perform edge-level prediction on the transaction graph. For both models, timestamps are converted into a scalar time feature by computing the number of seconds since midnight of the earliest day. For XGBoost, numerical features (time and monetary amounts, and the mediator variables in $T_E$) are used as-is since it is a tree-based method, and the categorical variables, including bank and account identifiers, currency types, payment formats, and, in the extended dataset, the sender and receiver countries, are represented via one-hot encoding. The model is trained for 300 boosting rounds with a maximum tree depth of four and a learning rate of 0.05, using subsampling and column subsampling rates of 0.8. The GNN classifiers use the tuned hyper-parameters of the original benchmark~\cite{altman_realistic_2023}. Training runs for 100 epochs using mini-batch edge sampling, where edge batches are constructed by neighbour sampling with 100 neighbours per layer and a batch size of 8\,192 edges. For the basic dataset $T_B$, no node information is used, as in the original publication. For $T_E$, country and transaction behaviour are included as node features.

All classifiers are evaluated using precision ($\mathrm{Prec}_1$), recall ($\mathrm{Rec}_1$), and F1-score ($\mathrm{F1}_1$) for the minority class (positive value). The dataset is split into training (64\%), validation (19\%), and test (17\%) sets using a temporal day-boundary scheme that assigns entire days to each split, preserving the chronological order of transactions. The same partition is applied to all models so that XGBoost and the GNN classifiers are trained and evaluated on identical subsets of transactions. To mitigate the severe class imbalance, XGBoost up-weights the money laundering class in the loss by a factor equal to the ratio of negatives to positives in the training split, while the GNN classifiers use a weighted cross-entropy loss~\cite{altman_realistic_2023}. The validation set is used to tune the decision threshold, selecting the threshold that maximizes $\mathrm{F1}_1$. After training, the selected GNN checkpoint is used to produce class‑1 softmax scores on the validation and test edges, from which the tuned decision threshold is applied. All experiments are conducted on a workstation equipped with two Intel Xeon Gold 6326 CPUs and one NVIDIA GeForce RTX 3090 (24~GB VRAM). XGBoost training takes approximately 5~minutes on CPU; GNN-GIN training takes approximately 2~hours and GNN-PNA approximately 3~hours per run on the GPU. All test-set results are reported in Table~\ref{tab:F1}.

\begin{table}[htb!]
    \centering
    \caption{XGBoost and GNN performance for the test set
    on the basic transaction data $T_B$ and on the extended dataset
    $T_E$.}\label{tab:F1}
    \begin{tabular}{l l r r r r}
        \toprule
        \textbf{Classifier} & \textbf{Dataset} & $\mathrm{F1}_1$ & $\mathrm{Prec}_1$ & $\mathrm{Rec}_1$ \\
        \midrule
        XGBoost & $T_B$ & 0.15 & 0.11 & 0.27\\
        XGBoost & $T_E$ & 0.19 & 0.14 & 0.27\\
        \midrule
        GNN-GIN & $T_B$ & 0.20 & 0.15 & 0.28\\
        GNN-GIN & $T_E$ & 0.42 & 0.39 & 0.45\\
        \midrule
        GNN-PNA & $T_B$ & 0.56 & 0.62 & 0.51\\
        GNN-PNA & $T_E$ & 0.58 & 0.60 & 0.56\\
        \bottomrule
    \end{tabular}
\end{table}

The XGBoost algorithm shows poor performance on both datasets $T_B$ and $T_E$. The much higher recall compared to the precision indicates a high false positive rate. Adding country and mediator information increases $\mathrm{F1}_1$ from 0.15 to 0.19. This indicates that the extended features provide some additional discriminative signal, but the tabular model still struggles to recover positive cases reliably. GNN-GIN performs much better on the extended dataset $T_E$, with an increase in $\mathrm{F1}_1$ from 0.20 to 0.42. The baseline performance of GNN-GIN on $T_B$ is comparable to XGBoost, again with a high false positive rate. For GNN-PNA, $\mathrm{F1}_1$ improves from 0.56 on $T_B$ to 0.58 on $T_E$. This shows that the GNN-PNA already captures much of the relevant structure from the basic transaction graph, yet still leverages the extended country and mediator features, leading to a slight performance gain on top of a substantially higher baseline. Overall, these results indicate that all three classifiers extract meaningful signal from $T_E$, as reflected by consistent performance gains. These models are intended to be representative, reasonably strong baselines that might be employed in practice rather than fully optimized detectors for this simulated dataset.
\section{Counterfactual Fairness Analysis}\label{sec:fairness}
Let us adapt the general framework from Section~\ref{sec:background} to the synthetic AML setting introduced in Sections~\ref{sec:data} and~\ref{sec:algos}. The correspondences between the variables are as follows:
\begin{itemize}
\item The protected attribute $A$ represents the country associated with an account. For the GNN classifiers, $A$ is a node-level feature attached to each account node. For the XGBoost classifier, $A$ is defined as the sending country for each transaction.
\item The mediator $M$ corresponds to the transaction behaviour. For XGBoost, $M$ is the mediator from the sender account, for the GNNs, this acts at the node level as a single feature.
\item The remaining observed attributes are collected in the covariate vector $X$.
For the GNNs, the transaction attributes (timestamp, amount, currency, payment format) are encoded as edge-level features, while bank and account identities form the graph nodes. For XGBoost, $X$ includes all features in the tabular representation.
\item Finally, $\hat{Y}$ denotes the binary prediction produced by a trained AML classifier (XGBoost, GNN-GIN or GNN-PNA) based on $(A,M,X)$. Notice that this differs between the three algorithms, in particular, XGBoost has a high false positive rate.
\end{itemize}
The structural mechanism $f_{\hat{Y}}$ is given by the trained classifier. Once the model parameters are fixed, this is a deterministic function of the input features $(A,X,M)$. Since the structural mechanism for $f_M$ is not known, it is approximated by a tree-based regression model $\hat{f}_M$ that takes as input the concatenated feature vector $(A,X)$. This yields the fitted mediator
\[
\hat M
=
\hat f_M(A,X)\,,
\]
which is used in place of the unobserved structural mediator in all counterfactual evaluations for both classifiers. In practice, when setting $A \leftarrow a$, we replace $M$ by its counterfactual value only for transactions whose factual sending country differs from $a$, while we keep mediators for transactions with $A = a$ at their factual values. We assess the quality of the learned mediator function $\hat f_M$ on the validation set and perform hyper-parameter tuning, using 50 trees, a maximum depth of 10, and a minimum leaf size of 10 in the final specification. The resulting mediator model achieves an $R^2$ of 0.80 with a mean absolute error of 1.87, indicating that $\hat f_M$ captures a substantial fraction of the variability in sender activity. We therefore treat $\hat f_M$ as a reasonable approximation of the structural mediator in the subsequent counterfactual fairness analysis.

For the GNN-based classifiers, our mediation framework requires additional assumptions. When constructing counterfactual worlds, we duplicate the entire transaction graph and replace all occurrences of the relevant sending country with the intervention value $A=a$, so the question we address is closer to \textit{what if this and all other transactions from this country had $a$} than \textit{what if only this single transaction had $a$}. Furthermore, because the information in a GNN is propagated through the graph structure, intervening on $A$ could, in principle, also change the neighbourhood context. This implementation, however, only updates the protected attribute on the affected transactions (and on other transactions from the same country) while keeping the remaining graph structure factual.

We now turn to counterfactual fairness diagnostics, introduced in Section~\ref{sec:background}. In the following, we set the reference value $a'$ to the United States (US), as this is the largest country class in the dataset. We first inspect the counterfactual flip rate, defined in Equation~\eqref{eq:FR}. This quantity measures the proportion of transactions whose predicted label changes when we intervene on the sending country, and we use it primarily to assess whether some countries are more unstable under the intervention $A \leftarrow \mathrm{US}$. Importantly, the flip rate is agnostic to the direction of change and therefore does not indicate whether a model is biased in favour of or against a given group.

\begin{figure}[htbp]
    \centering
    \caption{Counterfactual flip rates $\mathrm{FR}_{\hat Y}(\mathrm{US})$ under the intervention $A \leftarrow \mathrm{US}$, stratified by factual sending country and model (XGBoost, GNN-GIN, GNN-PNA). Results are based on simulated data and are not expected to reflect country-specific patterns.}
    \Description{Bar chart of counterfactual flip rates under the intervention $A \leftarrow \mathrm{US}$, shown separately for XGBoost, GNN-GIN, and GNN-PNA. The x-axis lists major sending countries ordered by number of transactions. The y-axis shows the percentage of transactions whose predicted label changes when the sender country is set to the US. All flip rates are below 1.5\%, with XGBoost generally having higher flip rates than the two GNN models and some countries, such as Saudi Arabia and Brazil, exhibiting notably higher instability than others.}
    \label{fig:flip_rates_by_country}
    \includegraphics[width=0.9\linewidth]{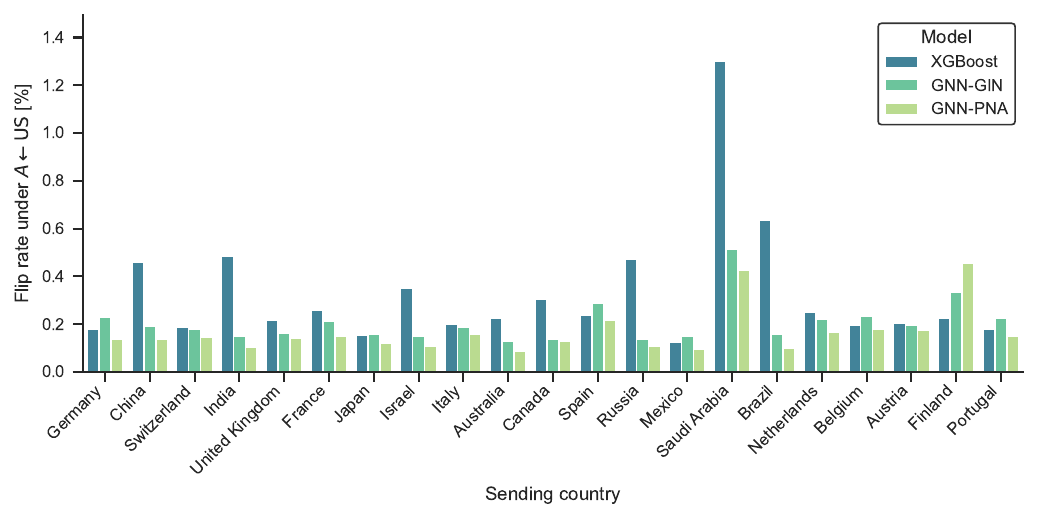}
\end{figure}

For XGBoost, the overall flip rate is 0.0015, meaning that 0.15\% of transactions change their predicted label when the sender country is counterfactually set to the US. The corresponding flip rates for the GNNs are similarly small, for GNN-GIN (0.0021) and for GNN-PNA (0.0014). To obtain a more granular view, we compute flip rates by factual sending country with more than 30\,000 transactions, see Figure \ref{fig:flip_rates_by_country} for the details.

The flip rates for transactions already sent from the US under the intervention $A \leftarrow \mathrm{US}$ are not depicted. For XGBoost, this is exactly zero, as expected, since the factual and counterfactual feature vectors coincide. For the GNNs, we have small but non-zero flip rates for the US under the same intervention. This arises from the preprocessing pipeline: the global perturbation $A \leftarrow \mathrm{US}$ changes the empirical feature distribution on the graph, per-split z-normalization is recomputed on this new distribution, and the resulting slight changes in normalized inputs cause a small number of predictions to flip. The country-level breakdown reveals that, despite overall flip rates being very small, there is heterogeneity across sending countries. For many countries, all models exhibit flip rates well below 0.3\%, indicating highly stable predictions. However, a few countries show noticeably higher flip rates, particularly for XGBoost, suggesting that predictions for these senders are more sensitive.

We now turn to the total, direct, and indirect effects introduced in Eqs.~\eqref{eq:TE},~\eqref{eq:NDE}, and \eqref{eq:NIE}. As noted in Section~\ref{sec:data}, the three classifiers have different baseline flagging rates, and the overall prevalence of laundering is very low. On the absolute scale, this implies that the effects are numerically small and not directly comparable across models. To obtain quantities that are easier to interpret and comparable between classifiers, we therefore rescale all three effects by the expected flagging rate under the reference value, that is, by $\mathbb{E}\big[\hat Y_{A \leftarrow \mathrm{US}}\big]$, which is 0.248 for XGBoost, 0.093 for GNN-GIN, and 0.035 for GNN-PNA. This yields relative total, direct, and indirect effects on the baseline-risk scale, denoted by
$\mathrm{RTE}_{\hat Y}(a,a')$,
$\mathrm{RDE}_{\hat Y}(a,a')$, and
$\mathrm{RIE}_{\hat Y}(a,a')$,
which quantify how strongly moving from the reference group (US) to a given country changes the predicted risk relative to the baseline flagging rate of each model.

\begin{figure}[htbp]
    \centering
    \caption{Estimated relative total (RTE), direct (RDE), and indirect (RIE) effects with respect to the US, shown as kernel density estimates over the 21 sending countries with at least 30\,000 transactions. Results are based on simulated data and are not expected to reflect country-specific patterns.}
    \Description{Three stacked panels showing kernel density estimates of relative total effect, relative direct effect, and relative indirect effect for three models: XGBoost, GNN-GIN, and GNN-PNA. The x-axis shows relative effect size and the y-axis density. In the top panel (RTE), all three densities are centred near zero, with XGBoost and GNN-GIN having wider, more right-skewed distributions and GNN-PNA more concentrated.  In the middle panel (RDE), all models have narrow densities tightly concentrated around zero, indicating small direct effects. In the bottom panel (RIE), XGBoost shows a broader distribution of indirect effects, while both GNN models have sharp peaks close to zero, suggesting that most indirect effects through the mediator are small.}
    \label{fig:effect_densities_by_model}
    \includegraphics[width=0.95\linewidth]{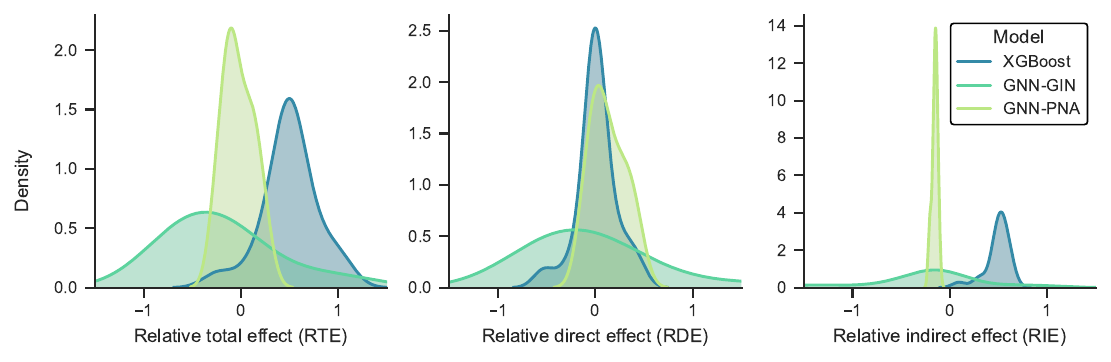}
\end{figure}

The resulting relative total, direct, and indirect effects for each country–model combination are summarized in Figure~\ref{fig:effect_densities_by_model}, which shows kernel density estimates of the three relative effects across the 21 sending countries considered. Numerical values for all country–model combinations aggregated in Figure~\ref{fig:effect_densities_by_model} are detailed in Table~\ref{tab:NDE_NIE_TE}.

\renewcommand{\arraystretch}{1.4}
\begin{table}[htb!]
\centering
\caption{Estimated relative total (RTE), direct (RDE), and indirect (RIE) effects with respect to the US,
by sending country and model, for all countries with more than
30\,000 transactions. 
Results are based on simulated data and are not expected to reflect country-specific patterns. Values correspond to the same estimates
summarized as densities in Figure~\ref{fig:effect_densities_by_model}.}\label{tab:NDE_NIE_TE}.
\begin{tabular}{lrrr|rrr|rrr}
\toprule
& \multicolumn{3}{c}{\textbf{XGBoost}} & \multicolumn{3}{c}{\textbf{GNN-GIN}} & \multicolumn{3}{c}{\textbf{GNN-PNA}} \\
\textbf{Country}
  & \multicolumn{1}{c}{$\mathrm{RTE}$}
  & \multicolumn{1}{c}{$\mathrm{RDE}$}
  & \multicolumn{1}{c}{$\mathrm{RIE}$}
  & \multicolumn{1}{c}{$\mathrm{RTE}$}
  & \multicolumn{1}{c}{$\mathrm{RDE}$}
  & \multicolumn{1}{c}{$\mathrm{RIE}$}
  & \multicolumn{1}{c}{$\mathrm{RTE}$}
  & \multicolumn{1}{c}{$\mathrm{RDE}$}
  & \multicolumn{1}{c}{$\mathrm{RIE}$}\\
\midrule
Australia & -0.24 & -0.53 & 0.29 & -0.79 & -0.65 & -0.14 & 0.03 & 0.19 & -0.16 \\
Austria & 0.50 & -0.01 & 0.51 & -0.73 & -0.52 & -0.21 & -0.27 & -0.14 & -0.12 \\
Belgium & 0.50 & -0.01 & 0.51 & 0.24 & 0.23 & 0.01 & 0.28 & 0.46 & -0.18 \\
Brazil & 0.68 & 0.12 & 0.56 & -0.17 & 0.39 & -0.56 & -0.05 & 0.10 & -0.16 \\
Canada & 0.48 & -0.01 & 0.49 & -0.85 & -0.79 & -0.05 & 0.13 & 0.34 & -0.21 \\
China & 0.46 & -0.01 & 0.47 & -0.45 & -0.31 & -0.14 & 0.02 & 0.16 & -0.14 \\
Finland & 0.55 & -0.02 & 0.57 & 0.06 & 0.88 & -0.82 & 0.13 & 0.33 & -0.20 \\
France & 0.49 & -0.01 & 0.50 & 0.29 & 1.71 & -1.41 & -0.12 & -0.00 & -0.12 \\
Germany & 0.47 & -0.02 & 0.49 & 0.98 & -0.03 & 1.01 & -0.21 & -0.09 & -0.12 \\
India & 1.06 & 0.43 & 0.63 & -0.77 & -0.52 & -0.25 & -0.18 & -0.04 & -0.14 \\
Israel & 0.30 & -0.15 & 0.45 & -0.07 & 0.28 & -0.35 & 0.18 & 0.33 & -0.16 \\
Italy & 0.49 & -0.01 & 0.50 & -0.54 & -0.41 & -0.13 & 0.08 & 0.25 & -0.17 \\
Japan & 0.72 & 0.16 & 0.57 & -0.55 & -0.50 & -0.04 & -0.06 & 0.09 & -0.16 \\
Mexico & 0.63 & 0.04 & 0.58 & -0.19 & -0.32 & 0.13 & -0.12 & 0.02 & -0.15 \\
Netherlands & 0.27 & -0.28 & 0.54 & -0.19 & -0.01 & -0.18 & -0.26 & -0.09 & -0.17 \\
Portugal & 0.50 & -0.01 & 0.51 & -0.26 & -0.14 & -0.11 & -0.13 & -0.03 & -0.10 \\
Russia & 0.82 & 0.23 & 0.59 & -0.37 & -0.15 & -0.22 & -0.12 & 0.03 & -0.15 \\
Saudi Arabia & 0.96 & 0.35 & 0.61 & 4.44 & 5.90 & -1.46 & 0.21 & 0.42 & -0.21 \\
Spain & 0.49 & -0.01 & 0.50 & 0.97 & 0.23 & 0.74 & -0.10 & 0.03 & -0.14 \\
Switzerland & 0.08 & -0.01 & 0.09 & -0.22 & 0.04 & -0.26 & -0.06 & 0.08 & -0.14 \\
United Kingdom & 0.31 & -0.01 & 0.32 & -0.59 & -0.49 & -0.10 & 0.09 & 0.22 & -0.14 \\
\midrule
Median & 0.49 & -0.01 & 0.51 & -0.22 & -0.14 & -0.14 & -0.06 & 0.09 & -0.15 \\
\bottomrule
\end{tabular}
\end{table}
\renewcommand{\arraystretch}{1}

Across all three models, these effects are small in absolute terms but non‑negligible when expressed relative to the US baseline flagging rate, indicating that the sending country protected attribute has a considerable causal influence on predictions beyond random noise. For XGBoost, the relative direct effects are tightly concentrated around zero, whereas the relative indirect effects show a clear positive mode, suggesting that the total effect of the sending country is moderate and predominantly mediated through the extended features rather than arising from a direct dependence on the country label itself. In contrast, GNN‑GIN exhibits larger and more dispersed relative direct effects, with large positive values for several countries and sometimes slightly negative indirect effects. Together with its strongest performance gains from the extended features, this suggests that GNN-GIN learns to exploit country-related information more aggressively than the other models, making it the least fair in our path-specific sense and illustrating a clear accuracy–fairness tension. Finally, GNN‑PNA shows both relative direct and indirect effects sharply concentrated near zero, in line with its low flagging rates and flip rates, indicating that it exploits the extended, country-related information only weakly and exhibits the smallest country-specific causal effects among the three classifiers.

The code used for the experiments together with the necessary preprocessing scripts discussed in this paper is freely available.\footnote{See \href{http://github.com/idsia-papers/2026-ecaf}{github.com/idsia-papers/2026-ecaf}.}
\section{Discussion and Conclusions}\label{sec:discussion}
We presented a counterfactual and path-specific fairness framework to analyse the potential impact of biases in machine learning–based predictive tools for AML. Building on a structural causal model for AML transaction monitoring, we analysed flip rates, total effects, and their decomposition into natural direct and indirect effects of three AML classifiers to quantify how changes in the protected feature would causally influence model outputs along pathways that are considered legitimate versus illegitimate. These approaches appear to provide a particularly informative lens for evaluating the influence of protected KYC features on AML predictions. Our experiments are based on simulated data, and it is important to emphasize that the country-specific results reported here are purely illustrative. They are not intended to reflect real-world statistics or money laundering patterns across jurisdictions.

We evaluated three representative AML classifiers to assess how they exploit the protected information. The relatively modest performance of XGBoost offers a realistic baseline for predictions achievable by a single financial institution relying solely on internal AML data. The two graph neural networks (GNN‑GIN and GNN‑PNA) leverage the transaction network structure to reach higher detection performance on the same task, highlighting the potential value of inter-institutional communication and data sharing. Such collaboration, however, would likely require the adoption of privacy-preserving techniques, such as federated learning~\cite{mcmahan_communication-efficient_2017}, to ensure confidentiality.

We find that the architecture that benefits most from the extended features in terms of $\mathrm{F1}_1$, GNN-GIN, is also the one that most strongly violates our counterfactual fairness criterion, illustrating a concrete accuracy–fairness trade off~\cite{plecko_fairness-accuracy_2025}. By contrast, the best-performing algorithm, GNN‑PNA, which did not improve performance by a lot with the extended features, only leads to small direct and indirect effects of the country attribute. For the tabular baseline, XGBoost, a large total effect is driven primarily through the mediated path, showing that relying solely on aggregate measures, such as the total effect, may obscure important mechanisms through which bias operates, and therefore fail to provide a complete picture.

Our findings build on several modelling assumptions that may affect the robustness of the analysis, particularly regarding extraction of the pseudo KYC features that act as a mediator and a protected feature. Because we work with simulated, open‑access data, both the protected attribute and the behaviour feature had to be constructed. In a real‑world deployment, the protected feature would typically be derived from KYC records, often in free‑text form. For simplicity, our analysis focuses on the origin of each transaction, using only the sending country and the mediator of the sending account, whereas extensions could incorporate receiving‑side information. A further limitation lies in the causal graph itself, in particular in how we distinguish the mediator from the remaining covariates. In practice, other features are likely to be causally influenced by the protected attribute, and alternative graph specifications could be considered. Finally, the choice of mediator model also influences the resulting effects. Here, we adopt a tree‑based regression as a flexible but pragmatic approximation of the mediator mechanism, leaving the exploration of alternative, e.g., neural, architectures for future work.

In contrast to the tabular setting, where causal mediation analysis is relatively well established, extending this framework to graph neural networks raises additional challenges. Fairness in GNNs is now an active research area~\cite{agarwal_towards_2021,ma_learning_2022,guo_towards_2023,chen_fairness-aware_2024}, with recent work focusing on bias mitigation~\cite{wang_advancing_2024,liu_fairness-aware_2026}. In our analysis, we adopted a deliberately simple intervention scheme for computational tractability. We duplicate the transaction graph and flip the protected attribute for the relevant nodes, while leaving the remaining graph structure and node attributes unchanged. This ignores potential feedback between the protected attribute of a node and its neighbourhood, but it provides a practical first step towards counterfactual fairness diagnostics for GNN‑based AML systems.

Overall, our findings underscore the importance of adopting nuanced fairness analyses in AML machine learning systems. In a setting with extreme class imbalance and substantial challenges in training effective machine learning models, there is pressure to focus on highly sensitive KYC information. Our results show that even in this simulated environment, a counterfactual fairness analysis can reveal when performance gains are driven by legitimate signal and when they depend on direct use of protected attributes, yielding a more complete view of fairness. We see this as a first step towards counterfactual fairness audits for AML models, tabular and graph‑based alike, which can help inform regulatory guidance on acceptable uses of sensitive KYC information in automated transaction monitoring.
\bibliographystyle{ACM-Reference-Format}
\bibliography{Bibtex_Fair_AML}

@article{nazar_magnitude_2024,
	title = {The magnitude and consequences of money laundering},
	volume = {27},
	issn = {1368-5201, 1368-5201},
	url = {http://www.emerald.com/jmlc/article/27/5/808-824/1225226},
	doi = {10.1108/jmlc-09-2022-0139},
	number = {5},
	journal = {Journal of Money Laundering Control},
	author = {Nazar, Sadia and Raheman, Abdul and Anwar Ul Haq, Muhammad},
	year = {2024},
	pages = {808--824},
}

@article{mazumder_explainable_2026,
	title = {Explainable and fair anti-money laundering models using a reproducible {SHAP} framework for financial institutions},
	volume = {6},
	issn = {2731-0809},
	url = {https://link.springer.com/10.1007/s44163-026-00944-7},
	doi = {10.1007/s44163-026-00944-7},
	number = {1},
	journal = {Discover Artificial Intelligence},
	author = {Mazumder, Pristly Turjo},
	year = {2026},
	pages = {262},
}

@book{pearl_causality_2009,
	address = {Cambridge},
	edition = {2},
	title = {Causality: models, reasoning, and inference},
	isbn = {978-0-521-89560-6},
	publisher = {Cambridge University Press},
	author = {Pearl, Judea},
	year = {2009},
}

@article{motie_financial_2024,
	title = {Financial fraud detection using graph neural networks: a systematic review},
	volume = {240},
	issn = {09574174},
	shorttitle = {Financial fraud detection using graph neural networks},
	url = {https://linkinghub.elsevier.com/retrieve/pii/S0957417423026581},
	doi = {10.1016/j.eswa.2023.122156},
	language = {en},
	journal = {Expert Systems with Applications},
	author = {Motie, Soroor and Raahemi, Bijan},
	year = {2024},
	pages = {122156},
}

@inproceedings{altman_realistic_2023,
	title = {Realistic synthetic financial transactions for anti-money laundering models},
	doi = {10.5555/3666122.3667422},
	booktitle = {Advances in {Neural} {Information} {Processing} {Systems} 36},
	publisher = {Curran Associates Inc.},
	author = {Altman, Erik and Blanuša, Jovan and von Niederhäusern, Luc and Egressy, Béni and Anghel, Andreea and Atasu, Kubilay},
	year = {2023},
	pages = {29851 -- 29874},
}

@inproceedings{gu_optimization_2025,
	title = {Optimization of anti-money laundering detection models based on causal reasoning and interpretable artificial intelligence and its empirical study on financial system stability},
	isbn = {979-8-4007-1951-6},
	url = {https://dl.acm.org/doi/10.1145/3785706.3785754},
	doi = {10.1145/3785706.3785754},
	booktitle = {Proceedings of the 2025 2nd {International} {Conference} on {Digital} {Economy} and {Computer} {Science}},
	publisher = {ACM},
	author = {Gu, Xiaoxiong and Yang, Jingwen and Liu, Min},
	year = {2025},
	pages = {304--308},
}

@inproceedings{kamalaruban_evaluating_2024,
	title = {Evaluating fairness in transaction fraud models: fairness metrics, bias audits, and challenges},
    isbn = {979-8-4007-1081-0},
	shorttitle = {Evaluating fairness in transaction fraud models},
	url = {https://dl.acm.org/doi/10.1145/3677052.3698666},
	doi = {10.1145/3677052.3698666},
	booktitle = {Proceedings of the 5th {ACM} {International} {Conference} on {AI} in {Finance}},
	publisher = {ACM},
	author = {Kamalaruban, Parameswaran and Pi, Yulu and Burrell, Stuart and Drage, Eleanor and Skalski, Piotr and Wong, Jason and Sutton, David},
	year = {2024},
	pages = {555--563},
}

@article{liu_fairness-aware_2026,
	title = {Fairness-aware graph representation learning through bias disentanglement},
    volume = {193},
	issn = {09505849},
	url = {https://linkinghub.elsevier.com/retrieve/pii/S0950584926000236},
	doi = {10.1016/j.infsof.2026.108034},
	journal = {Information and Software Technology},
	author = {Liu, Shuhan and Qin, Zheyun and Hou, Xuan and Wang, Yining and Wang, Ziwen and Peng, Zhaohui},
	year = {2026},
	pages = {108034},
}

@inproceedings{agarwal_towards_2021,
	title = {Towards a unified framework for fair and stable graph representation learning},
    volume = {161},
	url = {https://proceedings.mlr.press/v161/agarwal21b.html},
	doi = {10.48550/arXiv.2102.13186},
	booktitle = {Proceedings of {Machine} {Learning} {Research}},
	publisher = {PMLR},
	author = {Agarwal, Chirag and Lakkaraju, Himabindu and Zitnik, Marinka},
	year = {2021},
	pages = {2114--2124}}

@inproceedings{kusner_counterfactual_2017,
	title = {Counterfactual fairness},
	isbn = {978-1-5108-6096-4},
	doi = {10.5555/3294996.3295162},
	booktitle = {Proceedings of the 31st {International} {Conference} on {Neural} {Information} {Processing} {Systems}},
	publisher = {Curran Associates Inc.},
	author = {Kusner, Matt and Loftus, Joshua and Russell, Chris and Silva, Ricardo},
	year = {2017},
	pages = {4069--4079},
}

@inproceedings{ma_learning_2022,
	title = {Learning fair node representations with graph counterfactual fairness},
	isbn = {978-1-4503-9132-0},
	url = {https://dl.acm.org/doi/10.1145/3488560.3498391},
	doi = {10.1145/3488560.3498391},
	booktitle = {Proceedings of the {Fifteenth} {ACM} {International} {Conference} on {Web} {Search} and {Data} {Mining}},
	publisher = {ACM},
	author = {Ma, Jing and Guo, Ruocheng and Wan, Mengting and Yang, Longqi and Zhang, Aidong and Li, Jundong},
	month = feb,
	year = {2022},
	pages = {695--703},
}

@article{chen_fairness-aware_2024,
	title = {Fairness-aware graph neural networks: a survey},
	volume = {18},
	issn = {1556-4681, 1556-472X},
	shorttitle = {Fairness-aware graph neural networks},
	url = {https://dl.acm.org/doi/10.1145/3649142},
	doi = {10.1145/3649142},
	number = {6},
	journal = {ACM Transactions on Knowledge Discovery from Data},
	author = {Chen, April and Rossi, Ryan and Park, Namyong and Trivedi, Puja and Wang, Yu and Yu, Tong and Kim, Sungchul and Dernoncourt, Franck and Ahmed, Nesreen},
	year = {2024},
	pages = {1--23},
}

@inproceedings{guo_towards_2023,
	title = {Towards fair graph neural networks via graph counterfactual},
	isbn = {979-8-4007-0124-5},
	url = {https://dl.acm.org/doi/10.1145/3583780.3615092},
	doi = {10.1145/3583780.3615092},
	booktitle = {Proceedings of the 32nd {ACM} {International} {Conference} on {Information} and {Knowledge} {Management}},
	publisher = {ACM},
	author = {Guo, Zhimeng and Li, Jialiang and Xiao, Teng and Ma, Yao and Wang, Suhang},
	year = {2023},
	pages = {669--678},
}

@inproceedings{egressy_provably_2024,
	title = {Provably powerful graph neural networks for directed multigraphs},
	volume = {38},
	isbn = {978-1-57735-887-9},
	doi = {10.1609/aaai.v38i10.29069},
	booktitle = {Proceedings of the {AAAI} {Conference} on {Artificial} {Intelligence}},
	publisher = {AAAI Press},
	author = {Egressy, Béni and Von Niederhäusern, Luc and Blanuša, Jovan and Altman, Erik and Wattenhofer, Roger and Atasu, Kubilay},
	year = {2024},
	pages = {11838--11846},
}

@article{plecko_causal_2024,
	title = {Causal fairness analysis: a causal toolkit for fair machine learning},
	volume = {17},
	issn = {1935-8237, 1935-8245},
	shorttitle = {Causal fairness analysis},
	url = {https://www.emerald.com/ftmal/article/17/3/304/1332395/Causal-Fairness-Analysis-A-Causal-Toolkit-for-Fair},
	doi = {10.1561/2200000106},
	number = {3},
	journal = {Foundations and Trends in Machine Learning},
	author = {Plečko, Drago and Bareinboim, Elias},
	year = {2024},
	pages = {304--589},
}

@inproceedings{chiappa_path-specific_2019,
	title = {Path-specific counterfactual fairness},
	volume = {33},
	copyright = {https://www.aaai.org},
	issn = {2374-3468, 2159-5399},
	url = {https://ojs.aaai.org/index.php/AAAI/article/view/4777},
	doi = {10.1609/aaai.v33i01.33017801},
	booktitle = {Proceedings of the {AAAI} {Conference} on {Artificial} {Intelligence}},
	publisher = {AAAI Press},
	author = {Chiappa, Silvia},
	year = {2019},
	pages = {7801--7808},
}

@incollection{wang_advancing_2024,
	title = {Advancing graph counterfactual fairness through fair representation learning},
	volume = {14947},
	isbn = {978-3-031-70367-6 978-3-031-70368-3},
	url = {https://link.springer.com/10.1007/978-3-031-70368-3_3},
	doi = {10.1007/978-3-031-70368-3_3},
	language = {en},
	booktitle = {Machine {Learning} and {Knowledge} {Discovery} in {Databases}. {Research} {Track}},
	publisher = {Springer Nature Switzerland},
	author = {Wang, Zichong and Chu, Zhibo and Blanco, Ronald and Chen, Zhong and Chen, Shu-Ching and Zhang, Wenbin},
	year = {2024},
	pages = {40--58},
}

@inproceedings{mcmahan_communication-efficient_2017,
	title = {Communication-efficient learning of deep networks from decentralized data},
	volume = {54},
	url = {https://proceedings.mlr.press/v54/mcmahan17a.html},
	doi = {10.48550/arXiv.1602.05629},
	booktitle = {Proceedings of the 20th {International} {Conference} on {Artificial} {Intelligence} and {Statistics}},
	publisher = {Proceedings of Machine Learning Research},
	author = {McMahan, Brendan and Moore, Eider and Ramage, Daniel and Hampson, Seth and Arcas, Blaise Agüera},
	year = {2017},
	pages = {1273--1282},
}

@inproceedings{plecko_fairness-accuracy_2025,
	title = {Fairness-accuracy trade-offs: a causal perspective},
	volume = {39},
	issn = {2374-3468, 2159-5399},
	url = {https://ojs.aaai.org/index.php/AAAI/article/view/34833},
	doi = {10.1609/aaai.v39i25.34833},
	number = {25},
	booktitle = {Proceedings of the {AAAI} {Conference} on {Artificial} {Intelligence}},
	publisher = {AAAI Press},
	author = {Plečko, Drago and Bareinboim, Elias},
	year = {2025},
	pages = {26344--26353},
}
\end{document}